\documentclass[runningheads]{llncs}
\usepackage[T1]{fontenc}
\usepackage{graphicx}
\usepackage{subfigure}
\usepackage{amsmath}
\usepackage{amssymb}
\usepackage{booktabs}
\usepackage{colortbl}
\usepackage{multirow}
\usepackage{stfloats}
\usepackage{tikz,lscape,amsmath}
\usepackage{authblk}

\begin{document}
\title{GB-CosFace: Rethinking Softmax-based Face Recognition from the Perspective of Open Set Classification}
\titlerunning{GB-CosFace}
%
\author{Mingqiang Chen$\dagger$  \and
Lizhe Liu$\dagger$  \and
Xiaohao Chen \and
Siyu Zhu\inst{*}}
%
%
\institute{Alibaba Group \\
\email{mimingqiang.cmq, lizhe.llz, xiaohao.cxh, siting.zsy@alibaba-inc.com}}

\maketitle

\def\thefootnote{$\dagger$}\footnotetext{These authors contributed equally to this work}
\def\thefootnote{*}\footnotetext{Siyu Zhu is the corresponding author.}
\begin{abstract}
State-of-the-art face recognition methods typically take the multi-classification pipeline and adopt the softmax-based loss for optimization. Although these methods have achieved great success, the softmax-based loss has its limitation from the perspective of open set classification: the multi-classification objective in the training phase does not strictly match the objective of open set classification testing. In this paper, we derive a new loss named global boundary CosFace (GB-CosFace). Our GB-CosFace introduces an adaptive global boundary to determine whether two face samples belong to the same 
identity so that the optimization objective is aligned with the testing process from the perspective of open set classification. 
Meanwhile, since the loss formulation is derived from the softmax-based loss, our GB-CosFace retains the excellent properties of the softmax-based loss, and CosFace is proved to be a special case of the proposed loss. We analyze and explain the proposed GB-CosFace geometrically. Comprehensive experiments on multiple face recognition benchmarks indicate that the proposed GB-CosFace outperforms current state-of-the-art face recognition losses in mainstream face recognition tasks. Compared to CosFace, our GB-CosFace improves 
5.30\%, 0.70\%, and 0.36\% at TAR@FAR=1e-6, 1e-5, 1e-4 on IJB-C benchmark.

\end{abstract}
\section{Introduction}
Research on the training objectives of face recognition (FR) has effectively improved the performance of deep-learning-based face recognition\cite{wang2021deep,taigman2014deepface,sun2015deepid3,wen2016discriminative}. According to whether a proxy is used to represent a person's identity or a set of training samples, face recognition methods can be divided into proxy-free methods\cite{chopra2005learning,sun2015deep,ustinova2016learning,han2018face,schroff2015facenet,parkhi2015deep,ge2018deep,zhong2019adversarial,oh2016deep,rippel2015metric,sohn2016improved,wu2017sampling} and proxy-based methods\cite{sun2014deep,liu2017sphereface,wang2017normface,wang2018additive,wang2018cosface,sun2020circle,deng2019arcface,zheng2018ring,huang2020curricularface,meng2021magface,chen2017noisy,kim2020broadface,deng2021variational}. The proxy-free methods directly compress the intra-class distance and expand the inter-class distance based on pair-wise learning\cite{sun2015deep,ustinova2016learning,han2018face,chopra2005learning} or triplet learning\cite{schroff2015facenet,parkhi2015deep,ge2018deep,zhong2019adversarial,oh2016deep,wu2017sampling,sohn2016improved}. However, when dealing with a large amount of training data, the hard-mining operation which is crucial for proxy-free methods becomes extremely difficult. Recently, proxy-based method have achieved great success and shown advantages in big data training.
Most of them take a softmax-based multi-classification pipeline and use cross-entropy loss as the optimization objective. In these methods, each identity in the training set is represented by a prototype, which is the weight vector of the final fully connected layer. 
We refer to this type of method as the softmax-based face recognition method in this paper.
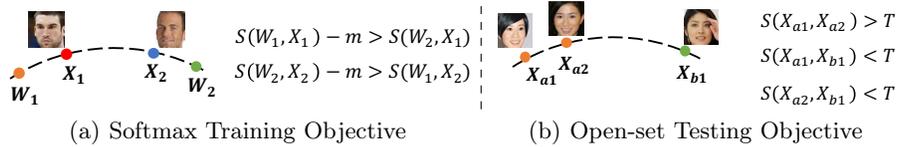
\begin{figure}[t]
\centering

\subfigure[Softmax Training Objective]{
\begin{minipage}[t]{0.52\linewidth}
\centering
\includegraphics[width=1\linewidth]{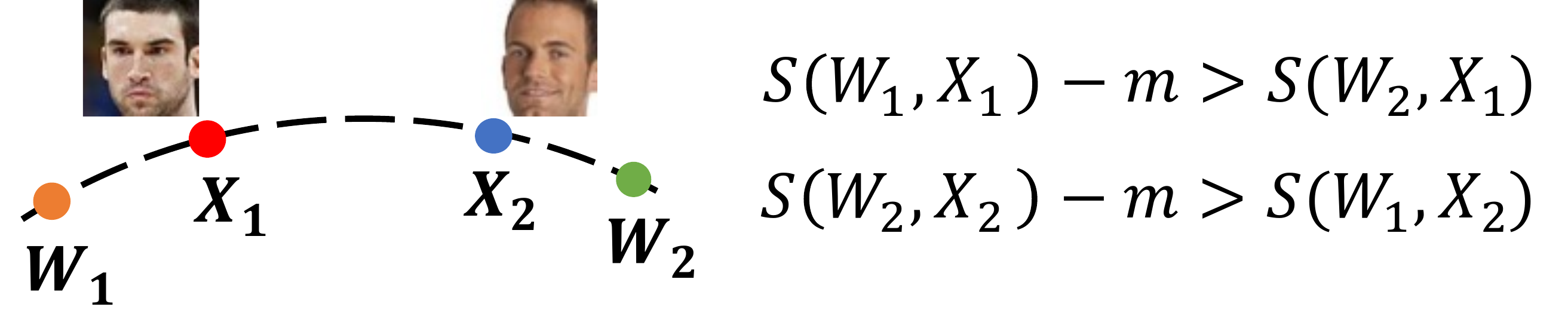}
\end{minipage}%
}%
\begin{tikzpicture}
\coordinate (a) at (0,0);
\coordinate (b) at (0,1.4);
\draw[black,dashed,thin] (a) -- (b);
\end{tikzpicture}
\subfigure[Open-set Testing Objective]{
\begin{minipage}[t]{0.44\linewidth}
\centering
\includegraphics[width=1\linewidth]{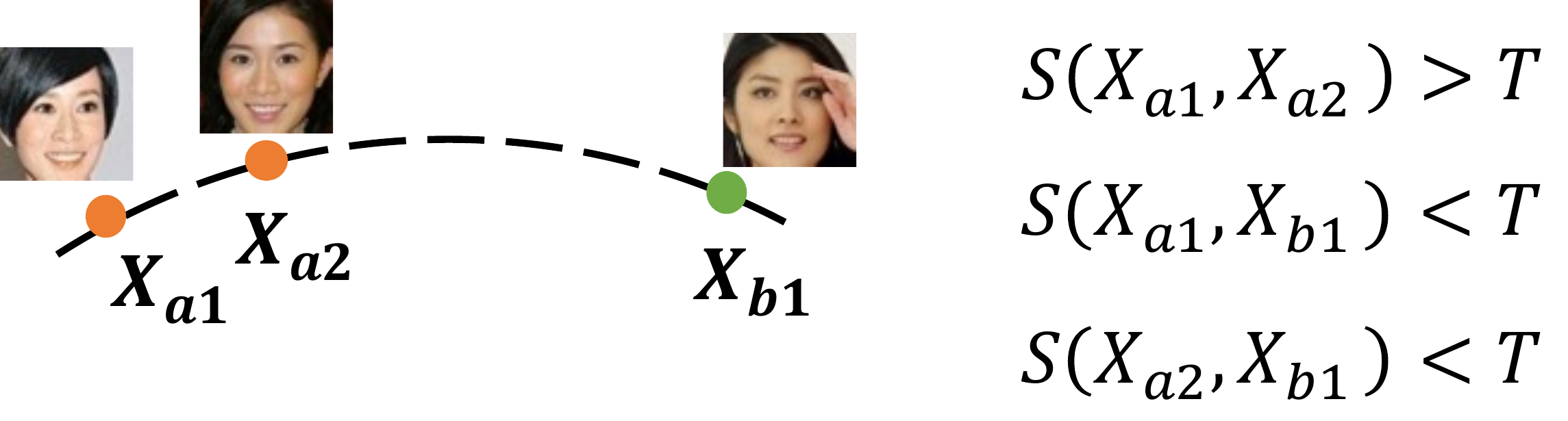}
\end{minipage}%
}%
\centering
\caption{The difference of the objective between softmax-based training and the open set classification testing, where \(S(\cdot)\) is the function to measure the distance between two samples, \(W_1\) and \(W_2\) are the prototypes of two identities respectively. In \textbf{Figure (a)}, \(X_1\) and \(X_2\) is the given training sample, \(m\) is the margin parameter.  In \textbf{Figure (b)}, \(X_{a1}\) and \(X_{a2}\) are two testing samples of ID ``a'', and \(X_{b1}\) is a testing sample of ID ``b''. ID ``a'' and ``b'' are not included in the training data.} 
\label{Fig.motivation} 
\end{figure}

Despite the great success of softmax-based face recognition, this strategy has its limitation from the perspective of the open set classification\cite{scheirer2012toward,geng2020recent,ge2017generative,yoshihashi2019classification}. As is shown in \textbf{Figure} \ref{Fig.motivation}(a), the training objective of softmax-based multi-classification is to make the predicted probability of the target category larger than other categories. However, face recognition is an open set classification problem where the test category generally does not exist in the training category~\cite{wang2021deep}. A typical requirement for a face recognition model is to determine whether two samples belong to the same identity by comparing the similarity between them with a global threshold \(T\), as is shown in \textbf{Figure} \ref{Fig.motivation}(b). The inconsistency of the objective of training and testing limits the performance. 

To reduce the impact of this inconsistency, current softmax-based face recognition methods have made various improvements to the training objective. 
One of the most vital improvements is to normalize the face features to the hyper-sphere for unified comparison~\cite{wang2017normface,liu2017sphereface}. Typically, the similarity between two samples is represented by the cosine similarity of their corresponding feature vectors. 
Large-margin-based methods\cite{liu2017sphereface,wang2018cosface,wang2018additive,deng2019arcface} are proposed to further compress the intra-class distance and expand the inter-class distance. Recently, the dynamic schemes for the scale parameter \cite{zhang2019adacos} and the margin parameter\cite{liu2019adaptiveface,meng2021magface} have been studied and further improved the model performance.

From the perspective of training strategy, Lu et al.\cite{lu2019sampling} proposed an optimal sampling strategy to address the inconsistency between the direction of gradient descent and optimizing the concerned evaluation metric. For face feature alignment, DAM\cite{liu2021dam} proposed a Discrepancy Alignment Metric, which introduces local inter-class differences for each face feature obtained from a pre-trained model, in the face verification stage. However, none of these methods consider introducing the global boundary in the testing process into the training objective.

In this paper, we propose a novel face recognition loss named global boundary CosFace (GB-CosFace), which resolve the above-mentioned inconsistencies well and can be easily applied for end-to-end training on face recognition task. In our GB-CosFace loss, the training objective is aligned with the testing process by introducing a global boundary determined by the proposed adaptive boundary strategy. First, we compare the objective difference between the softmax-based loss and the face recognition testing process. Then, we abstract the reasonable training objective from the perspective of open set classification and derive a antetype of the proposed loss. Furthermore, we combine the excellent properties of softmax-based losses with the proposed antetype loss and derive the final GB-CosFace formulation. We further prove that CosFace\cite{wang2018additive,wang2018cosface} is a special case of the proposed GB-CosFace. Finally, we analyze and explain the proposed GB-CosFace geometrically. The contributions of this paper are summarized as follows.

\begin{itemize}
    \item We propose GB-CosFace loss for face recognition, which matches the testing objective of the open set classification while inheriting the advantages of the softmax-based loss. To the best of our knowledge, we are the first work which introduces a global boundary into the training objective for face recognition.
    \item We analyze the difference and connection between GB-CosFace and general softmax-based losses, and give a reasonable geometric explanation.
    \item Our GB-CosFace obviously improve the performance of softmax-based face recognition (e.g.,
    improves 5.30\%, 0.70\%, and 0.36\% at TAR@FAR=1e-6, 1e-5, 1e-4 on IJB-C benchmark 
    compared to CosFace).

\end{itemize}
\section{Softmax-based Face Recognition}
To better understand the proposed GB-CosFace, this section review the general softmax-based face recognition.

\subsection{Framework}

\begin{figure}[t]
\centering
\includegraphics[width=0.9\linewidth]{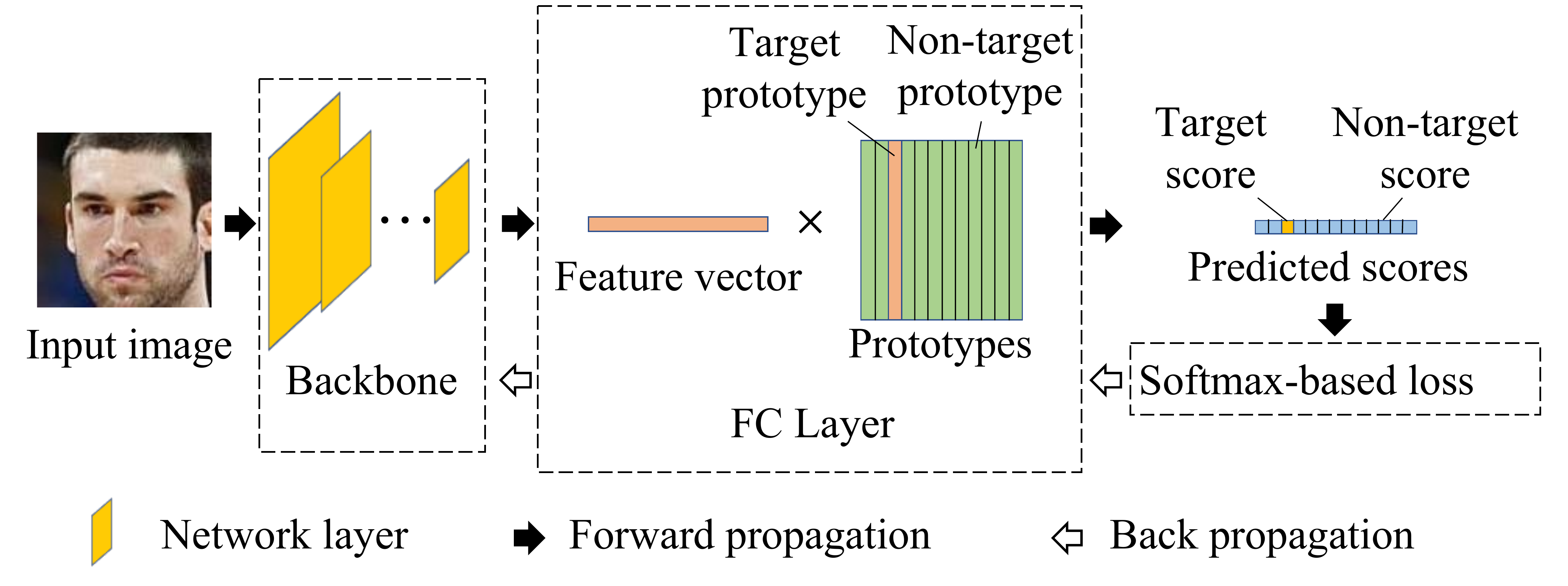}
\caption{The training framework of the general softmax-based face recognition.} 
\label{Fig.pipeline} 
\end{figure}
The training framework of the general softmax-based face recognition is shown in \textbf{Figure} \ref{Fig.pipeline}. In this framework, each identity in the training set has its corresponding prototype. The prototypes are represented by the weight vectors of the final fully connected layer. Given a training sample, we call the prototype representing the identity of this sample ``target prototype'', and call other prototypes ``non-target prototypes''. After extracting face features from the backbone, the predicted scores which represent the similarity between the feature vector and each prototype are calculated through the final fully connected layer (FC layer). The similarity between the feature vector and the target prototype is called ``target score'', and the other predicted scores are called ``non-target scores''. 
Generally, the output feature vector and the prototypes are normalized to the unit hyper-sphere. Therefore, the predicted scores are usually represented by the cosine of the feature vector and the prototype. 
In training, the softmax-based loss is adopted to optimize the backbone and the final FC layer through backpropagation.

\subsection{Objective}
\label{sec:objective}

For each iteration in n-class face recognition training, given a training sample and its label \(y\), the general softmax-based loss is as follows:

\begin{equation}
    \displaystyle
    {\mathcal{L}_S=-log\frac{e^{s{(cos(\theta _y +m_\theta )}-m_p)}}{ {\textstyle e^{s(cos(\theta _y +m_\theta )-m_p)}+\sum_{i}e^{s{cos_{\theta i}}}}}}
    \label{Eq.L1} 
\end{equation}%
where \(\theta_y\) is the arc between the predicted feature vector and the target prototype, \(\theta_i\) is the arc between the predicted feature vector and the non-target prototype, \(y\) is the index of the target identity, \(i\) is the index of the non-target identities, \(i \in [1,n]\) and \(i\ne y\).
There are three hyper-parameters: the scale parameter ``\(s\)'', and the two margin parameters ``\(m_\theta\)'' and ``\(m_p\)''.  

We can reach several common softmax-based losses from \textbf{Equ.} \ref{Eq.L1}. E.g., normalized softmax loss will be reached if both \(m_\theta\) and \(m_p\) are set as zero. ArcFace and CosFace will be reached if we respectively set \(m_p\) and \(m_\theta\) as 0.

Softmax-based losses can be regarded as the smooth form of the following optimization objective \(\mathcal{O}_{S}\).

\begin{equation}
    \displaystyle
    \begin{aligned}
      \mathcal{O}_{S}&= ReLU(max(cos\theta_i )-(cos(\theta_y+m_\theta  )-m_p)) \\
        &= \lim_{s \to +\infty}-\frac{1}{s}log\frac{e^{s(cos(\theta_y+m_\theta)-m_p)}}{e^{s(cos(\theta_y+m_\theta)-m_p)}+\sum_{i=1,i\ne y}^{n}e^{scos\theta_i}} \\ 
        &= \lim_{s \to +\infty}\frac{1}{s} \mathcal{L}_S
    \end{aligned} 
    \label{Eq.Os}
\end{equation}%
Where the SoftPlus function is used as a smooth form of ReLU operator and \(log\sum exp(\cdot )\)  is used as a smooth form of \(max(\cdot)\) operator. More detailed
derivation is included in the supplementary material.


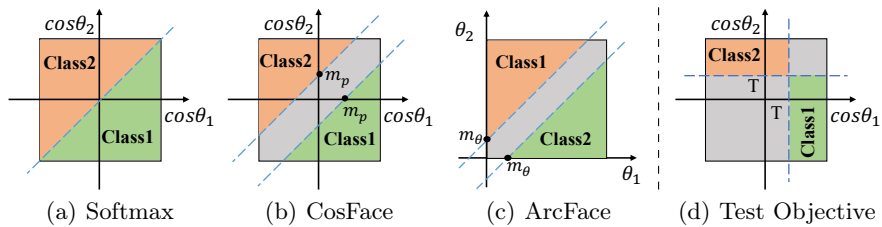
\begin{figure}[t]
\centering

\subfigure[Softmax]{
\begin{minipage}[t]{0.23\linewidth}
\centering
\includegraphics[width=1\linewidth]{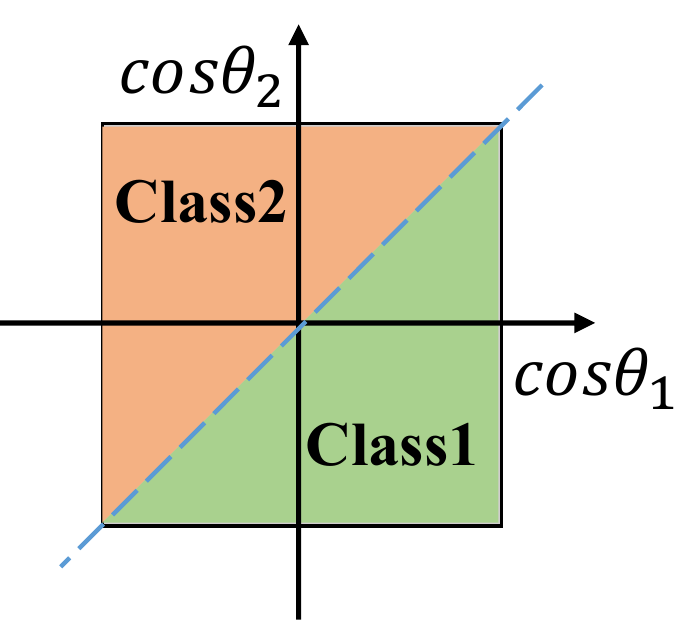}
\end{minipage}%
}%
\subfigure[CosFace]{
\begin{minipage}[t]{0.23\linewidth}
\centering
\includegraphics[width=1\linewidth]{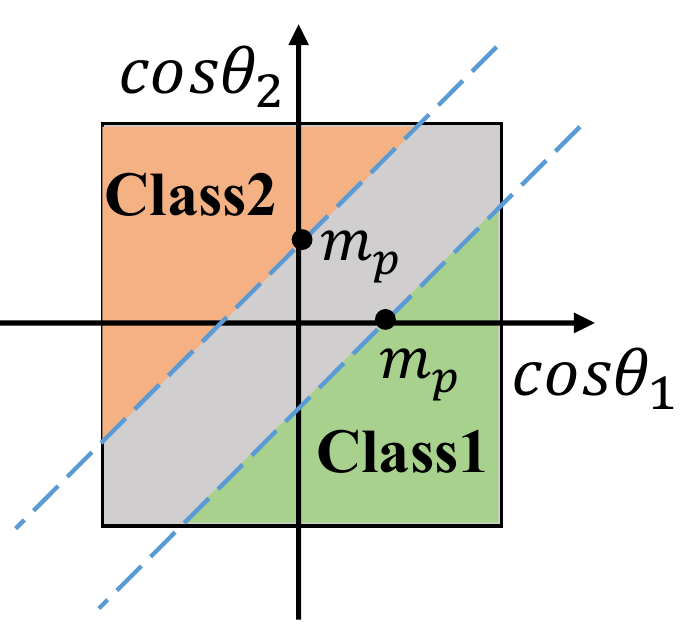}
\end{minipage}%
}%
\subfigure[ArcFace]{
\begin{minipage}[t]{0.23\linewidth}
\centering
\includegraphics[width=1\linewidth]{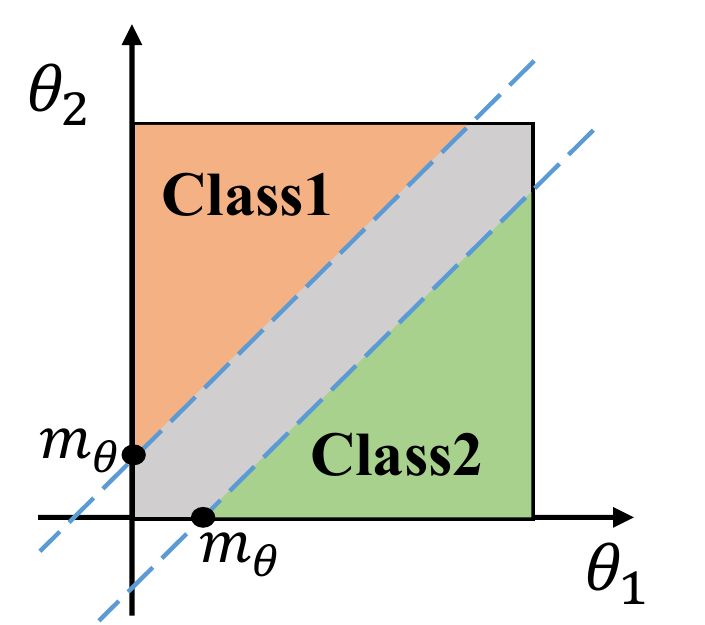}
\end{minipage}%
}%
\begin{tikzpicture}
\coordinate (a) at (0,0);
\coordinate (b) at (0,2.5);
\draw[black,dashed,thin] (a) -- (b);
\end{tikzpicture}
\subfigure[Test Objective]{
\begin{minipage}[t]{0.23\linewidth}
\centering
\includegraphics[width=1\linewidth]{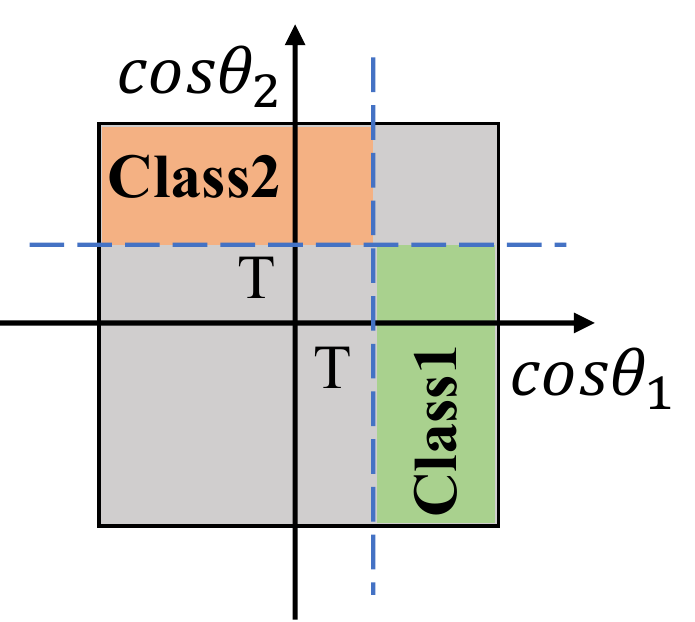}
\end{minipage}%
}%
\caption{Decision boundaries of different loss functions under binary
classification case. Figure (d) shows the expected decision boundary in the testing phase. } 
\label{Fig.divide} 
\end{figure}

From this perspective, we can find that the training objective \(\mathcal{O}_{S}\) constrains the target score to be larger than the maximum non-target score. The margin is introduced for a stricter constraints.
However, this constraint is not completely consistent with the objective of the testing process. Based on \textbf{Equ.} \ref{Eq.Os}, we can visualize the decision boundaries of normalized softmax loss\cite{wang2017normface}, CosFace\cite{wang2018cosface,wang2018additive}, and ArcFace\cite{deng2019arcface} under binary classification case, as is shown in \textbf{Figure} \ref{Fig.divide} (a)-(c).  In the testing phase, a global threshold \(T\) of the cosine similarity needs to be fixed to determine whether two samples belong to the same person, as is shown in \textbf{Figure} \ref{Fig.divide}(d). We can see that, even if a margin is added, the decision boundaries of softmax-based losses do not completely match the expected boundary for testing.

\subsection{Properties}
\label{sec:property}

Current face recognition models do not directly apply \(\mathcal{O}_{S}\) as the training objective. On the one hand, \(max(\cdot)\) operator only focuses on the maximum value and the gradients will only be backpropagated to the target score and maximum non-target score. On the other hand, if the argument of the RELU function is less than 0, no gradient will be backpropagated. As a smooth form of \(\mathcal{O}_{S}\), the softmax-based loss can avoid the above problems. The success of softmax-based loss is due to its excellent properties.

\textbf{Property 1.} The gradients of the non-target scores are proportional to their softmax value.

For softmax-based loss, the backpropagated gradients will be assigned to all non-target scores according to their softmax value. This property ensures that each non-target prototype can play a role in training, and hard non-target prototypes get more attention.

\textbf{Property 2.} The gradient of the target score and the sum of the gradients of all non-target scores have the same absolute value and opposite signs.

\begin{equation}
    \displaystyle
      {\frac{\partial \mathcal{L}_S}{\partial cos(\theta _y+m_\theta )}=-{\textstyle \sum_{i}} \frac{\partial \mathcal{L}_S}{\partial cos(\theta _i)}} \\
    \label{Eq.p2}
\end{equation}%

Softmax-based loss has balanced gradients for the target score and the non-target scores. This property can maintain the stability of training and prevent the training process from falling into a local minimum.

Considering the key role that these two properties play in face recognition training, we expect to inherit them in the loss design. In this paper, we add the consistency of training and testing to the loss design by introducing an adaptive global boundary. From the expected training objective, we derive our GB-CosFace framework and prove compatibility with CosFace. This compatibility allows the proposed loss to inherit the excellent properties of the general softmax-based loss while solving the inconsistency between the training and testing objective.

\section{GB-CosFace Framework}
\label{sec:method}
\subsection{Antetype Formulation}
Based on the face recognition testing process which is shown in \textbf{Fig} \ref{Fig.divide}(d), we propose to introduce a global threshold \(p_v\) as the boundary between target score and non-target scores. The target score is required to be larger than \(p_v\) while the maximum of the non-target scores is required to be less than \(p_v\). Following this idea, we improve \textbf{Equ.} \ref{Eq.Os} as follows:

\begin{equation}
    \displaystyle
    \left\{\begin{array}{l}
    \mathcal{O}_{T}=ReLU(p_v-(p_y-m))
    \\
    \mathcal{O}_{N}=ReLU(max(p_i)-(p_v-m))
    \end{array}\right.
    \label{Eq.etn}
\end{equation}%
where we divide the training objective into the target score \(\mathcal{O}_{T}\) and the non-target scores \(\mathcal{O}_{N}\) respectively. \(p_y\) is the target score, where \(p_y=cos\theta_y\). \(p_i\) is the non-target score, where \(p_i=cos\theta_i\). \(m\) is the margin parameter introduced for stricter constraints. The training objective is to minimize \(\mathcal{O}_{T}\) and \(\mathcal{O}_{N}\).

Inspired by the success of the softmax-based loss, similar to \textbf{Equ.} \ref{Eq.Os}, we take the smooth form of \(\mathcal{O}_{T}\) and \(\mathcal{O}_{N}\) as the antetype of the proposed loss.

\begin{equation}
    \displaystyle
    \left\{\begin{array}{l}
    \mathcal{L}_{T1}=-log\frac{e^{s(p_y-m)}}{e^{s(p_y-m)}+e^{sp_v}} 
    \\
    \mathcal{L}_{N1}=-log\frac{e^{s(p_v-m)}}{e^{s(p_v-m)}+ {\textstyle \sum_{i}} e^{sp_i}} 
    \end{array}\right.
    \label{Eq.base}
\end{equation}%
The loss for target score and non-target scores are represented as  \( \mathcal{L}_{T1}\) and \(\mathcal{L}_{N1}\) respectively. \(p_v\) is the global boundary hyper-parameter, which also means ``virtual score''. For \(\mathcal{L}_{T1}\), \(p_v\) is a virtual non-target score. For \(\mathcal{L}_{N1}\), \(p_v\) is a virtual target score. Since we take \(log\sum exp(\cdot )\) as the smooth form of \(max(\cdot)\), the distribution of the gradients of non-target scores inherits \textbf{Property 1.} (stated in \textbf{Section} \ref{sec:property}) of the softmax-based loss. 

\begin{figure}[hpbt]
\centering
\includegraphics[width=0.7\linewidth]{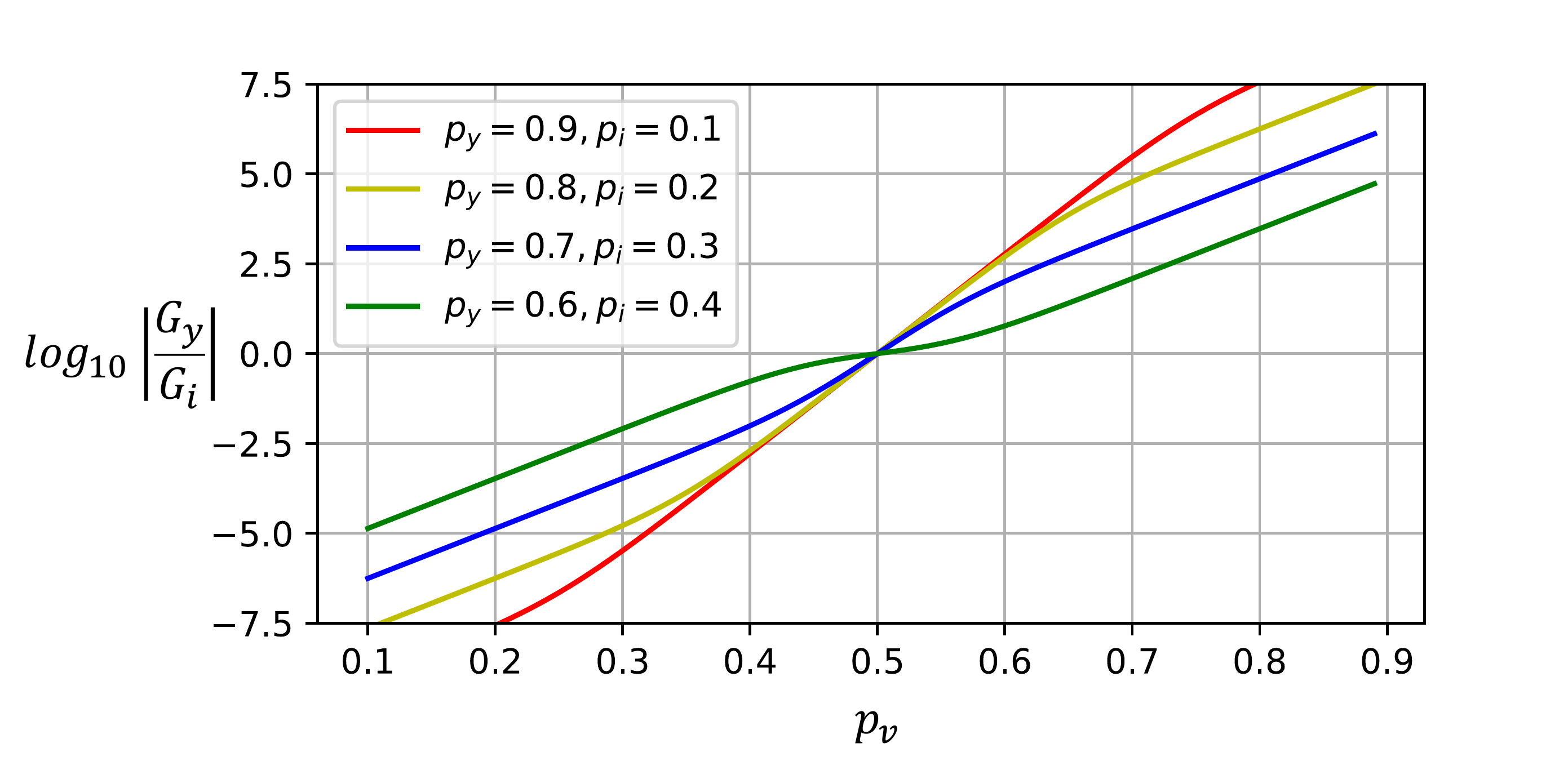}
\caption{The ratio of the target gradient to the non-target gradient varies with \(p_v\) under binary classification case using different \(p_y\) and \(p_i\).  Hyper-parameter \(s\) and \(m\) are set to 32 and 0.15 respectively. Note that the ordinate is the base 10 logarithm of the ratio.} 
\label{Fig.pv} 
\end{figure}

However, the proposed antetype introduces another problem: the setting of hyper-parameter \(p_v\).
First, the inappropriate setting of \(p_v\) may cause a serious gradient imbalance problem. Since we separate the constraints on the target score and the non-target scores, the gradient balance for target score and non-target scores is broken and the antetype loss no longer retains \textbf{Property 2.} (stated in \textbf{Section} \ref{sec:property}). Second, considering the rapid rise of the exponential function and the amplification effect of the hyper-parameter ``s'', the model is extremely sensitive to the choice of the hyper-parameter \(p_v\). As can be seen in \textbf{Figure} \ref{Fig.pv}, a slight change in \(p_v\) can cause an order of magnitude difference between the gradients for target score \(p_y\) and non-target scores \(p_i\). 
Therefore, an adaptive scheme for the global boundary is necessary.

\subsection{Adaptive Global Boundary}
\label{sec:adaptive-boundary}
To control the gradient balance and adapt the global boundary to different training stages, we propose an adaptive global boundary method. We believe that an ideal global boundary should meet the following conditions: \textbf{a)} Under this boundary setting, the gradients of the target score and the non-target scores should be roughly balanced from a global perspective; \textbf{b)} The global boundary should change slowly during the training process to keep the training stable while adapting to different training stages. Based on these two conditions, we make the following design.

\subsubsection{Gradient Balance Control.}

We define \(\hat{p}_{v}\) as the balanced threshold of the target score and the non-target scores which satisfies \(\frac{\partial \mathcal{L}_{T1}}{\partial p_{y}} =-  {\textstyle \sum_{i}} \frac{\partial \mathcal{L}_{N1}}{\partial p_{i}}\). Based on this condition, we reach the following form of \(\hat{p}_{v}\):
\begin{equation}
    \displaystyle
    {\hat{p}_{v}= (p_{y}+\frac{1}{s}log {\textstyle \sum_{i}}  e^{sp_{i}})/2}
    \label{Eq.pvj}
\end{equation}%

Ideally, for each iteration, to satisfy the above condition \textbf{a)}, we expect to calculate \(\hat{p}_v\) for each sample in the data set and get the mean value as the threshold \(p_v\). Considering the efficiency, we calculate the mean of \(\hat{p}_v\) for each batch and update it by the momentum update strategy.

\begin{equation}
    \displaystyle
    {p_{vg}=(1-\gamma)p_{vg}+\gamma p_{vb}}
    \label{Eq.vg}
\end{equation}%
Where \(\gamma \in [0,1]\) is the update rate, \(p_{vb}\) is the mean of \(p_{v}\) in a batch. A small \(\gamma\) can keep the stability of \(p_v\). We empirically set \(\gamma\) to 0.01.

This dynamic threshold strategy makes the gradient balanced globally. However, for each sample, the problem of gradient imbalance can be very serious.  Therefore, we modify the value of \(p_v\) to be the weighted sum of \(p_{vg}\) and \(\hat{p}_v\) as follows.

\begin{equation}
    \displaystyle
    {p_v=\alpha p_{vg}+(1-\alpha)\hat{p}_v}
    \label{Eq.alpha}
\end{equation}%

Where \(\alpha\) is a hyper-parameter and \(\alpha\in [0,1]\). When \(\alpha=0\), the gradients for the target score and the non-target scores are completely balanced. We can control the degree of the gradient imbalance by adjusting \(\alpha\).

\subsubsection{Compatible with CosFace.}

In \textbf{Equ.} \ref{Eq.alpha}, if we take \(\alpha\) as \(0\), the proposed loss will fully conform to \textbf{Property 1} and \textbf{Property 2} (stated in \textbf{Section} \ref{sec:property}) of softmax-based loss. Through the following analysis, we can further find that Cosface\cite{wang2018cosface,wang2018additive}  is a special case of the proposed loss when \(\alpha=0\).

The gradients based on CosFace is calculated as follows.
\begin{equation}
    \displaystyle
    \begin{aligned}
      \mathcal{G}_{T-CosFace} &= -\mathcal{G}_{N-CosFace}
        =-\frac{s\cdot {\textstyle \sum_{i}} e^{s{p_i}}}{e^{s{(p_y-m)}}+{\textstyle \sum_{i}} e^{s{p_i}}}
        = -\frac{s\cdot e^{sp_n}}{e^{s(p_y-m)}+e^{sp_n}} \\
    \end{aligned} 
    \label{Eq.g_cosface}
\end{equation}%
Where the gradient for the target score is represented as \( \mathcal{G}_{T-CosFace}\), the sum of the gradients of the non-target scores is represented as \(\mathcal{G}_{N-CosFace}\), and \(p_n=\frac{1}{s} log {\textstyle \sum_{i}^{}e^{sp_i}} \).

For the proposed loss, based on \textbf{Equ.} \ref{Eq.base}, we can get the gradient for target score \(p_y\) (\( \mathcal{G}_{T1}\)) and the sum of the gradients for non-target scores \(p_i\) (\(\mathcal{G}_{N1}\)) when \(\alpha\) is set to 0.
\begin{equation}
    \displaystyle
    \begin{aligned}
      {\mathcal{G}_{T1}=-\mathcal{G}_{N1}=-\frac{s\cdot e^{\frac{1}{2} sp_n}}{e^{\frac{1}{2}s(p_y-2m)}+e^{\frac{1}{2}sp_n}}} \\
    \end{aligned} 
    \label{Eq.g_l1v}
\end{equation}%

As the above equation shows, if we take \(p_v\) as \(\hat{p}_v\) (\textbf{Equ.} \ref{Eq.pvj}), the difference of the proposed loss (\textbf{Equ.} \ref{Eq.base}) and CosFace only lies on the margin and the scale. The more detailed proof is included in the supplementary material.

\subsubsection{Final Loss.}

For formal unity with CosFace, we rewrite the proposed loss into the following form.

\begin{equation}
    \displaystyle
    \begin{aligned}
      \mathcal{L}_{GB-CosFace} &= -\frac{1}{2} log\frac{e^{2s(p_y-m)}}{e^{2s(p_y-m)}+e^{2sp_v}}
       -\frac{1}{2}log\frac{e^{2s(p_v-m)}}{e^{2s(p_v-m)}+ e^{2sp_n}} \\
    \end{aligned} 
    \label{Eq.final}
\end{equation}%
Where \(p_{n}=\frac{1}{s}log {\textstyle \sum_{i}}  e^{sp_{i}}\). The value of \(p_v\) is in accordance with \textbf{Equ.} \ref{Eq.alpha}. In training, \(p_v\) is a detached parameter which does not require gradients.

This is the final form of the proposed GB-CosFace. Under this formulation, the hyper-parameter \(\alpha\) controls the degree of gradient imbalance. If we set \(\alpha\) as 0, the gradients for the target score and the non-target scores are balanced, and the proposed GB-CosFace is equivalent to CosFace which has the margin of \(2m\) and the scale of \(s\).

\subsection{Geometric Analysis}
\label{sec:analysis}

To analyze the properties of the proposed loss and compare it with other softmax-based losses, we analyze the loss boundaries in the binary classification case. The boundaries of ArcFace\cite{deng2019arcface} and CosFace\cite{wang2018cosface,wang2018additive} are determined by the following \textbf{Equ.} \ref{Eq.arc_boundary} and \textbf{Equ.} \ref{Eq.cos_boundary} respectively.

\begin{equation}
    \displaystyle
      |arccos(P\cdot P_1)-arccos(P\cdot P_2)|=m \\
    \label{Eq.arc_boundary}
\end{equation}%

\begin{equation}
    \displaystyle
      |P\cdot P_1-P\cdot P_2|=m \\
    \label{Eq.cos_boundary}
\end{equation}%
Where \(P\) is the predicted normalized \(n\)-dimensional feature vector and \(n\) is the face feature dimension, \(P_1\) and \(P_2\) are the feature vectors of ID1 and ID2 respectively. 

\begin{figure}[h]
\centering
\includegraphics[scale=0.3]{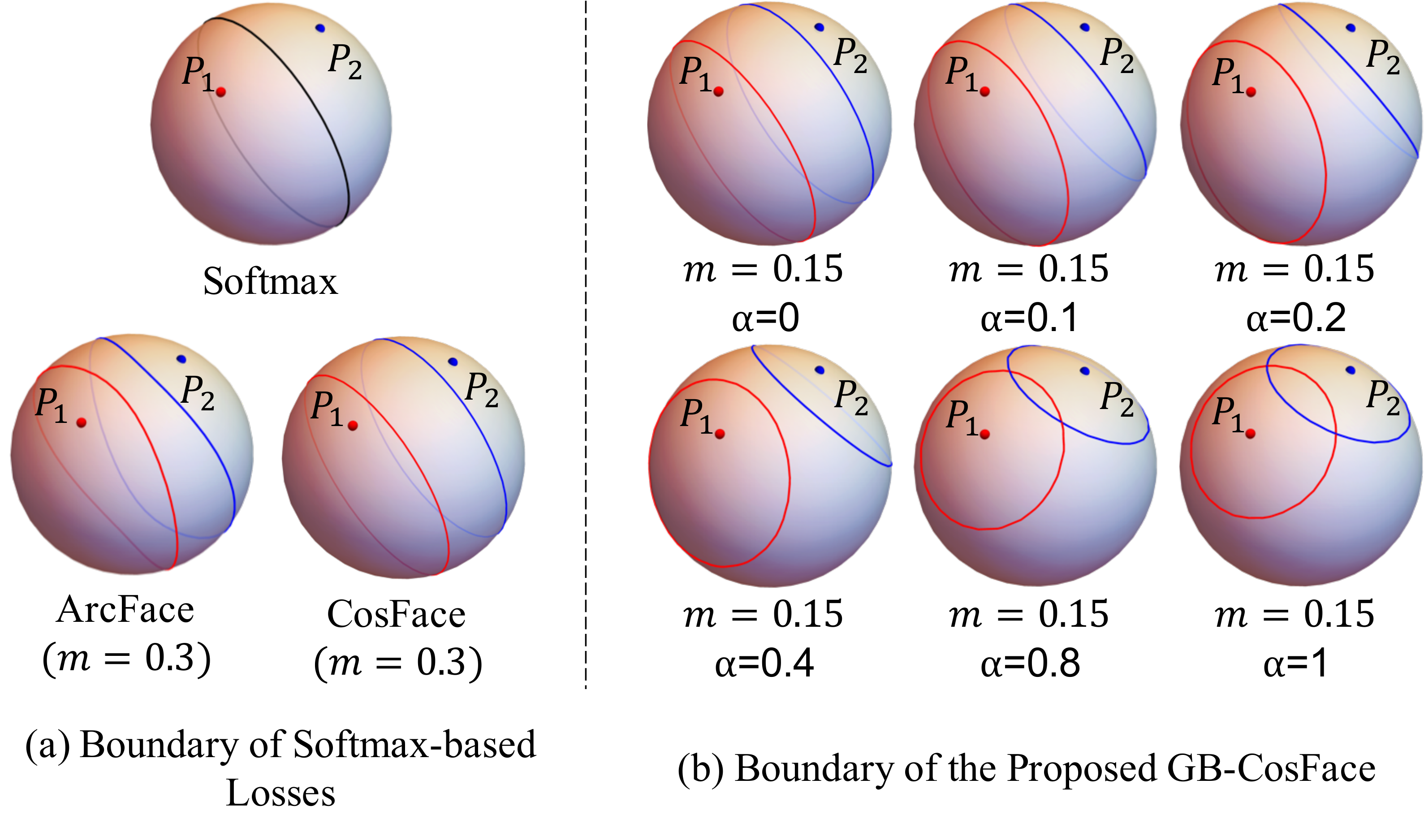}
\caption{Boundaries of the softmax-based losses and the proposed GB-CosFace loss. \(P_1\) and \(P_2\) are two points at a distance of \(60^\circ\). The red line and blue line are the target boundaries for \(P_1\) and \(P_2\) respectively. For the normalized softmax loss, the boundaries for \(P_1\) and \(P_2\) are coincident and represented in black color.
} 
\label{Fig.sphere} 
\end{figure}

For normalized softmax loss, the boundary is determined by \textbf{Equ.} \ref{Eq.arc_boundary} or \textbf{Equ.} \ref{Eq.cos_boundary} with a zero margin. We set the angle between vector \(P_1\) and \(P_2\) as \(60^\circ\) and show the boundaries of normalized softmax, ArcFace, and CosFace in the 3D spherical feature space in \textbf{Figure} \ref{Fig.sphere}(a). 

The boundaries of the proposed GB-CosFace can be determined according to \textbf{Equ.} \ref{Eq.ab_pos}.

\begin{equation}
    \displaystyle
    \left\{\begin{array}{l}
    |P\cdot P_1|=p_v+m 
    \\
    |P\cdot P_2|=p_v+m 
    \end{array}\right.
    \label{Eq.ab_pos}
\end{equation}%

According to \textbf{Equ.} \ref{Eq.pvj} and \textbf{Equ.} \ref{Eq.alpha}, in the binary classification case, \(p_v\) can be represented as follows.

\begin{equation}
    \displaystyle
      p_v=\alpha p_{vg}+(1-\alpha )(P\cdot P_1+P\cdot P_2)/2 \\
    \label{Eq.pv_sup}
\end{equation}%

We show the boundaries of the proposed GB-CosFace loss in \textbf{Figure} \ref{Fig.sphere}(b), where \(p_{vg}\) is fixed to 0.62 (a reasonable value according to the experiments in \textbf{Section} \ref{sec:experiments}) and \(m\) is fixed to 0.15.

In the face recognition problem, feature vectors of the same identity are expected to cluster together. However, by observing \textbf{Figure} \ref{Fig.sphere}(a), we can find that the boundaries in the case of binary classification do not meet this expectation. Only the positions near the line from point \(P_1\) to point \(P_2\) on the sphere can be effectively constrained. Fortunately, the training set has far more than two identities. Ideally, the prototypes of different identities will be evenly distributed on the sphere. The feature vectors of the same identity will be constrained in all directions. But actually, it cannot be guaranteed that in the sparse high-dimensional spherical feature space, there are enough non-target prototypes evenly distributed around each training sample.

The proposed loss \(\mathcal{L}_{GB}\) alleviates this problem by introducing a global boundary. As is shown in \textbf{Figure} \ref{Fig.sphere}(b), when \(\alpha=0\), the boundary is the same as CosFace. When \(\alpha=1\), the boundary is a circle on the sphere centered on \(P_1\) or \(P_2\) with a fixed radius completely determined by \(p_{vg}\) and the margin \(m\). With the increase of \(\alpha\), the boundary is closer to the ideal open set classification objective. However, an excessively large \(\alpha\) will cause blurring or even crossing of the boundaries between different identities.

{\begin{figure*}[h]
\centering
\includegraphics[scale=0.35]{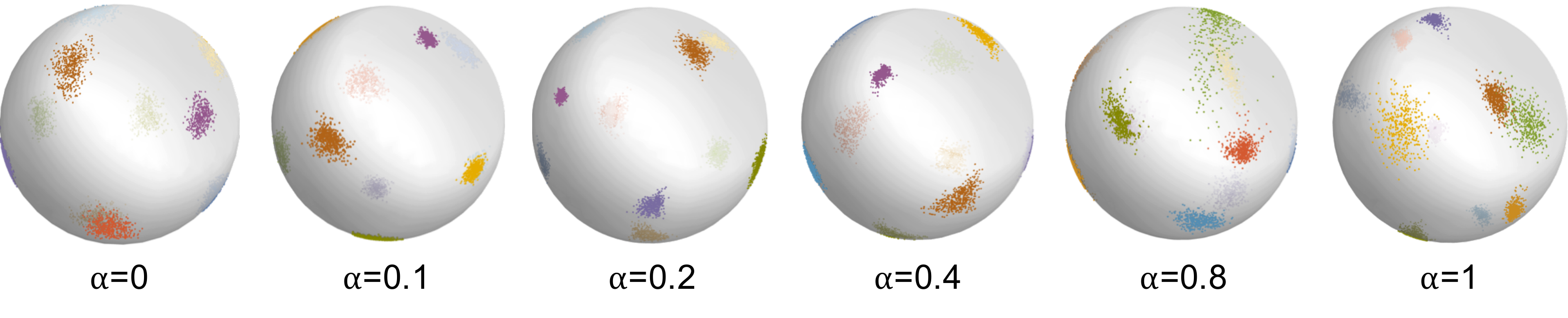}
\caption{Visualization of the toy experiments on the proposed GB-CosFace. Different colors represent different identities.}
\label{Fig.toy} 
\end{figure*}}

To study the appropriate range of \(\alpha\), we conduct a toy experiment based on a seven-layer convolutional neural network on a small face recognition dataset containing ten identities. We set the feature dimension as three, and visualize the distribution of the feature vectors on the unit sphere under different \(\alpha\) settings, as is shown in \textbf{Figure} \ref{Fig.toy}. The margin is fixed to 0.15 and \(\alpha\) is adjusted from 0 to 1. When \(\alpha=0\), our GB-CosFace is exactly the same as CosFace with the margin of 0.3, as indicated in \textbf{Section} \ref{sec:adaptive-boundary}. As \(\alpha\) increases, e.g., \(\alpha=0.2\), the feature vectors of the same identity are more concentrated as expected. The model performance will deteriorate if \(\alpha\) is further increased, e.g., \(\alpha=0.8\) or \(\alpha=1\). The setting of \(\alpha\) will be studied in detail in the \textbf{Section} \ref{sec:ablation}.

\section{Experiments}
\label{sec:experiments}
In this section, we verify our GB-CosFace on two important face tasks: face recognition and face clustering. Furthermore, we conduct ablation experiments to verify the proposed strategies and the settings of the hyper-parameters.

\subsubsection{Dataset.} 
We employ MS1MV3\cite{deng2019lightweight}, a refined version of MS1M\cite{guo2016ms} as our training set for all the following experiments. This is a large-scale face recognition dataset containing 5.1M face images of 93K celebrities. We use several popular benchmarks as the validation set, including LFW\cite{huang2008labeled}, CFP-FP\cite{sengupta2016frontal}, CPLFW\cite{zheng2018cross}, AgeDB-30\cite{moschoglou2017agedb}, and CALFW\cite{zheng2017cross}. And we use  IJB-B\cite{whitelam2017iarpa} and IJB-C\cite{maze2018iarpa} as the testing sets. 

\subsubsection{Implementation Details.}
We use ResNet50\cite{he2016deep} and ResNet100 as the backbone for the following experiments. The BN-FC-BN structure is added after the last convolution layer to output 512-dimensional face feature vectors. For data pre-possessing, all face images are set to $112 \times 112$ and normalized by utilizing five facial points following recent papers\cite{deng2019arcface,meng2021magface}. Each RGB pixel is normalized to \([-1,1]\). Random horizontal flip is the only data augmentation method employed in the training process.
For optimization, we adopt the stochastic gradient descent (SGD) optimizer with a momentum of 0.9 and weight decay of 1e-4. 
We adopt the step learning rate decay strategy with an initial learning rate of 0.1. We train 24 epochs and divide the learning rate by 10 at 5, 10, 15, and 20 epochs. 
The training batch size is fixed to 512. Eight NVIDIA GPUS are employed for training.
We fix the hyper-parameters \(s\), \(m\), \(\alpha\) and \(\gamma\) as 32, 0.16, 0.15 and 0.01 respectively if not specified.

\subsection{Face Recognition}
\label{sec:exp-rec}
\begin{figure}[t]
\centering
\subfigure[Gradients during training.]{
\begin{minipage}[t]{0.5\linewidth}
\centering
\includegraphics[width=1\linewidth]{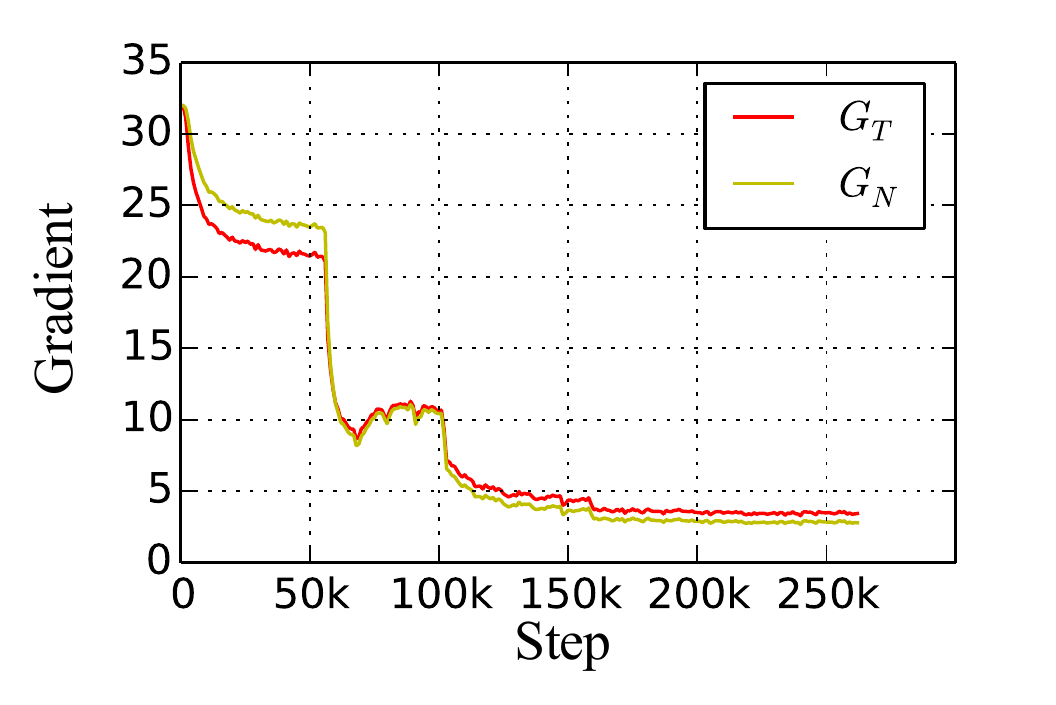}
\end{minipage}%
}%
\subfigure[Global boundary \(p_v\) during training.]{
\begin{minipage}[t]{0.5\linewidth}
\centering
\includegraphics[width=1\linewidth]{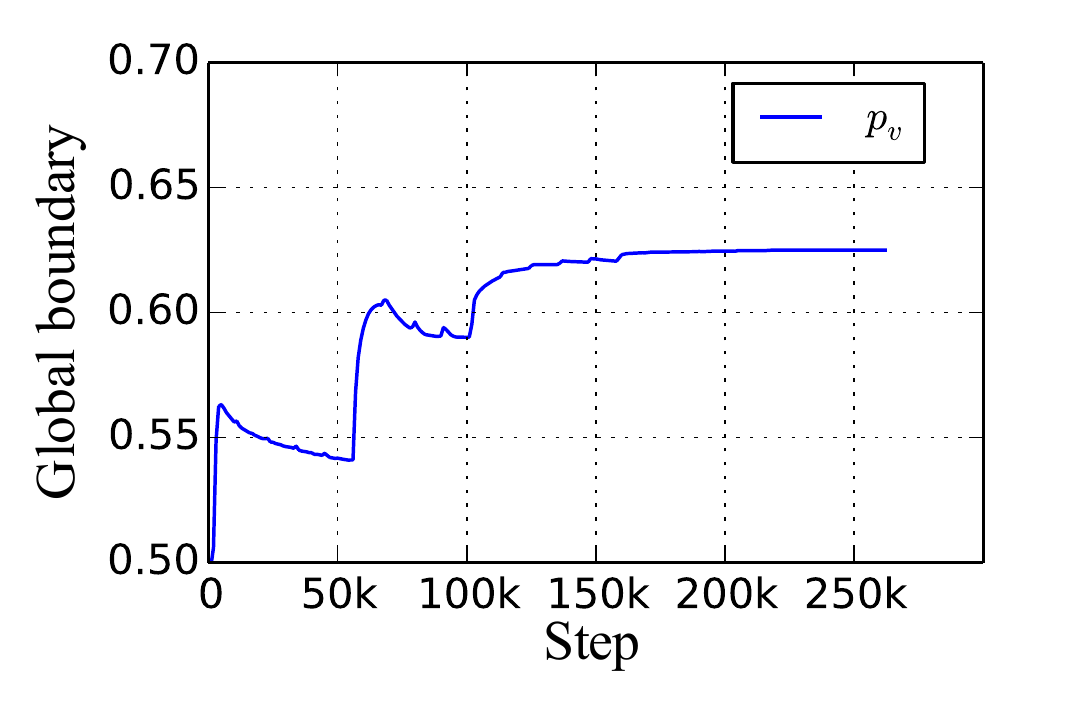}
\end{minipage}%
}%
\centering
\caption{\(G_T\) is the gradient of the target score, \(G_N\) is the gradients sum of the non-target scores, and \(p_v\) is the global boundary in \textbf{Equ.} \ref{Eq.final}.} 
\label{Fig.gradient} 
\end{figure}

\subsubsection{Analysis of Gradient Blance.}
\textbf{Figure} \ref{Fig.gradient} shows the gradients and the global boundary \(p_v\) in the training process. Throughout the training process, the gradient of the target score \(G_T\) and the gradients sum of the non-target scores \(G_N\) maintain a same convergence trend, and the values of \(G_T\) and \(G_N\) are approximately equal after 50k iterations. The change trend of the global boundary parameter \(p_v\) during the training process is consistent with the gradients \(G_T\) and \(G_N\), and eventually converges to 0.62. This result shows that our adaptive global boundary strategy can guarantee the stability of model training and keep \(G_T\) and \(G_N\) balanced throughout the training process, which is consistent with our discussion in \textbf{Section} \ref{sec:adaptive-boundary}.

\subsubsection{Results on Validation Datasets.}

To compare with recent state-of-the-art competitors, we compare the results on several popular face recognition benchmarks, including LFW, CFP-FP, AgeDB-30, CALFW, and CPLFW. LFW focuses on unconstrained fa
ce verification. 

\begin{table}[!hbp]
    \centering
    \caption{1:1 verification accuracy is reported on the LFW, CFP-FP, AgeDB-30, CALFW, CPLFW datasets. Backbone network: ResNet100.}
    \setlength{\tabcolsep}{3.5pt}{
    \scalebox{1}{
    \begin{tabular}{l|ccccc}
    \hline 
    \hline
    \multirow{2}{0cm}{\centering
    Method}&\multicolumn{5}{c}{Validation Dataset}\\
    \cline{2-6} &LFW&CFP-FP&AgeDB-30&CALFW&CPLFW\\
    \hline
    CosFace\cite{wang2018cosface} (CVPR18)& 99.81 & 98.12 & 98.11 & 95.76 & 92.28\\ 
    ArcFace\cite{deng2019arcface} (CVPR19)& \textbf{99.83} & 98.27 & 98.28 & 95.45 & 92.08\\
    Sub-center ArcFace\cite{deng2020sub} (ECCV20)& \textbf{99.83} & 98.80 & \textbf{98.45} & - & - \\
    BroadFace\cite{kim2020broadface} (ECCV20)& \textbf{99.83} & 98.63 & 98.38 & \textbf{96.20} & 93.17 \\ 
    CurricularFace\cite{huang2020curricularface} (CVPR20)& 99.80 & 98.37 & 98.32 & \textbf{96.20} & 93.13\\
    URFace\cite{shi2020towards} (CVPR20)& 99.78 & 98.64 & - & - & -\\
    CosFace+SCF\cite{li2021spherical} (CVPR21)& 99.80 & 98.59 & 98.26 & 96.18 & 93.26\\
    MagFace\cite{meng2021magface} (CVPR21)& \textbf{99.83} & 98.46 & 98.17 & 96.15 & 92.87\\
    \hline 
    GB-CosFace & 99.80  & \textbf{98.84} & 98.31 & 96.15 & \textbf{93.55} \\
    \hline
    \end{tabular}
    }
    }
    \label{tab:lfw}
\end{table}
The results are shown in \textbf{Table} \ref{tab:lfw}. We achieve the best results on two of the five benchmarks. Even though both datasets are highly-saturated, our GB-CosFace surpasses the recent methods on CFP-FP and  CPLFW, and achieves comparable results on other three datasets.

\subsubsection{Results on IJB-B and IJB-C.}
IJB is one of the largest and most challenging benchmarks to evaluate unconstrained face recognition. IJB-B contains 1845 identities with 55025 frames and 7011 videos. IJB-C is an extension of IJB-B which contains about 3.5K identities from 138K face images and 11K face videos. 

\begin{table}[hbpt]
    \centering
    \caption{The face verification accuracy on IJB-B and IJB-C. We evaluated the TAR@FAR from 1e-4 to 1e-6. Backbone network: ResNet100.}
    \setlength{\tabcolsep}{4pt}{
    \scalebox{1.0}{
    \begin{tabular}{l|ccc|ccc}
    \hline 
    \hline
    \multirow{2}{0cm}{\centering
    Method}&\multicolumn{3}{c|}{IJB-B(TAR)}&\multicolumn{3}{c}{IJB-C(TAR)}\\
    \cline{2-7} &1e-6&1e-5&1e-4&1e-6&1e-5&1e-4\\
    \hline
    CosFace\cite{wang2018cosface} (CVPR18)          & 36.49 & 88.11 & 94.80 & 85.91 & 94.10 & 96.37 \\ 
    ArcFace\cite{deng2019arcface} (CVPR19)          & 38.28 & 89.33 & 94.25 & 89.06 & 93.94 & 96.03 \\
    Sub-center ArcFace\cite{deng2020sub} (ECCV20)   & - & - & \textbf{95.25} & - & - & 96.61 \\
    BroadFace\cite{kim2020broadface} (ECCV20)           & 46.53 & 90.81 & 94.61 & 90.41 & 94.11 & 96.03\\ 
    CurricularFace\cite{huang2020curricularface} (CVPR20) & - & - & 94.80 & - & - & 96.10  \\
    GroupFace\cite{kim2020groupface} (CVPR20)& \textbf{52.12} & 91.24 & 94.93 & 89.28 & 94.53 & 96.26 \\
    CosFace+DAM\cite{liu2021dam} (ICCV21)& - & - & 94.97 & - & - & 96.45\\ 
    CosFace+SCF\cite{li2021spherical} (CVPR21)& - & 91.02 & 94.95 & - & 94.78 & 96.22\\ 
    MagFace\cite{meng2021magface} (CVPR21) & 40.91 & 89.88 & 94.33 & 89.26 & 93.67 & 95.81 \\
    \hline 
    GB-CosFace & 42.56 & \textbf{92.18} & 95.21 & \textbf{91.21} & \textbf{94.80} & \textbf{96.73}\\
    \hline
    \end{tabular}
    }
    }
    \label{tab:ijb}
\end{table}

The results are shown in \textbf{Table} \ref{tab:ijb}. We achieve SOTA results on IJB-B and IJB-C. Compared to CosFace, our GB-CosFace improves 6.07\%, 4.07\% and 0.41\% at TAR@FAR=1e-6, 1e-5, 1e-4 on IJB-B, and improves 5.30\%, 0.70\% and 0.36\% at TAR@FAR=1e-6, 1e-5, 1e-4 on IJB-C.

\subsection{Ablation Study}
\label{sec:ablation}

To analyze the effect of the adaptive boundary strategy and the setting of hyper-parameter \(\alpha\), we train ResNet-50 networks on MS1MV3 with different settings and evaluated the TAR@FAR=1e-4 on IJB-C.

\subsubsection{Hyperparameter Setting.}
Compared to CosFace, we introduce another hyper-parameter \(\alpha\) in \textbf{Equ.}\ref{Eq.alpha}.
Since the settings of the scale parameter \(s\) and the margin parameter \(m\) have been studied in detail in the previous works\cite{wang2018cosface,wang2018additive,deng2019arcface}, we empirically set \(s=32\) and \(m=0.16\) (equivalent to \(m=0.32\) in CosFace), and focus on the setting of \(\alpha\).
For more detailed theoretical analysis, please refer to \textbf{Section} \ref{sec:analysis}.

\begin{table}[hbpt]
\begin{center}
    \caption{The results of the proposed GB-CosFace under different settings of \(\alpha\).}
    \label{tab:alpha}
    \setlength{\tabcolsep}{6pt}{
    \centering
    \begin{tabular}{l|c}
    \hline
    {\centering Settings}&IJB-C(TAR)\\
    \hline 
     FAR=1e-4, R50, adaptive \(p_v\), $\alpha$=0    & 96.10\\
     FAR=1e-4, R50, adaptive \(p_v\), $\alpha$=0.05 & 96.15 \\
     FAR=1e-4, R50, adaptive \(p_v\), $\alpha$=0.15 & 96.24 \\
     FAR=1e-4, R50, adaptive \(p_v\), $\alpha$=0.25 & \textbf{96.35} \\
     FAR=1e-4, R50, adaptive \(p_v\), $\alpha$=0.35 & 96.33 \\
     FAR=1e-4, R50, adaptive \(p_v\), $\alpha$=0.60 & 96.08 \\
    \hline
    \end{tabular}}
    \end{center}
\end{table}

We conduct the controlled experiment where the value of \(\alpha\) is set from 0 to 0.6 and other  parameters are fixed. 
The results are shown in \textbf{Table} \ref{tab:alpha}. When the value of  \(\alpha\) gradually increases from 0 to 0.25, the performance of the model gradually improves and the model performs best with \(\alpha=0.25\). After the value of  \(\alpha\) exceeds 0.35, the model performance obviously degenerates with the increase of \(\alpha\). Overall, the performance of the model can maintain relatively good results as the value of \(\alpha\) is between 0.15 and 0.35. This result is consistent with the previous discussion and the toy experiments in \textbf{Section} \ref{sec:analysis}.

\subsubsection{Effect of the Adaptive Boundary Strategy.}
To evaluate the effectiveness of the adaptive boundary strategy, we compare the fixed boundary strategy and the proposed adaptive boundary strategy in \textbf{Section} \ref{sec:adaptive-boundary}. We fix the \(p_v\) in our GB-CosFace(\textbf{Equ.}\ref{Eq.final}) to different values and keep the other experimental settings the same as \textbf{Section} \ref{sec:exp-rec}. Since \(p_v\) converges to 0.62 in the experiment in \textbf{Section} \ref{sec:exp-rec}, we choose \(p_v=0.62\) and additionally choose values near 0.62.

\begin{table}[hbpt]
\begin{center}

    \caption{Comparison of the results of the proposed adaptive global boundary strategy and the fixed global boundary strategy.}
    \label{tab:adaptive}
    \setlength{\tabcolsep}{6pt}{
    \centering
    \scalebox{1}{
    
    \begin{tabular}{l|c}
    \hline 
    {\centering Settings} &IJB-C(TAR@FAR=1e-4)\\
    \hline
    FAR=1e-4, R50, \(\alpha\)=0.15, \(p_v\)=0.50 & 91.19\\
    FAR=1e-4, R50, \(\alpha\)=0.15, \(p_v\)=0.58 & \textbf{96.27}\\
    FAR=1e-4, R50, \(\alpha\)=0.15, \(p_v\)=0.62 & 96.19\\
    FAR=1e-4, R50, \(\alpha\)=0.15, \(p_v\)=0.66 & 96.17\\
    FAR=1e-4, R50, \(\alpha\)=0.15, \(p_v\)=0.74 & 95.09\\
    \hline 
    FAR=1e-4, R50, \(\alpha\)=0.15, adaptive \(p_v\) & 96.24\\
    
    \hline
    \end{tabular}}
    }
    \end{center}
\end{table}

The results are shown in \textbf{Table} \ref{tab:adaptive}. For the fixed boundaries, the model performs best when \(p_v=0.58\) and gets worse rapidly when the \(p_v\) changes, eg. the TAR decreases to 91.19\% when \(p_v=0.50\). What's more, if we reduce \(p_v\) to 0.42 or increase it to 0.82, the training will not converge. For the adaptive boundary strategy, the performance is very close to the best fixed boundary strategy result. This indicates that for the fixed boundary strategy, the model performance is sensitive to the value of \(p_v\), a very careful setting of \(p_v\) is required to obtain good results. While the adaptive global boundary strategy does not require careful tuning of hyper-parameters to achieve a similar performance. This result is consistent with the previous discussion and the toy experiments in \textbf{Section} \ref{sec:analysis}.

\section{Conclusion}

In this work, we discuss the inconsistency between the training objective of the softmax-based loss and the testing process of face recognition, and derive a new loss from the perspective of open set classification, called the global boundary CosFace(GB-CosFace). Our GB-CosFace aligns the training objective with the face recognition testing process while inheriting the good properties of the softmax-based loss. 
Comprehensive experiments indicate that our GB-CosFace has an obvious improvement over general softmax-based losses.
%
%
%

\bibliographystyle{splncs}
\bibliography{egbib}

\begin{thebibliography}{10}

\bibitem{wang2021deep}
Wang, M., Deng, W.:
\newblock Deep face recognition: A survey.
\newblock Neurocomputing (2021)

\bibitem{taigman2014deepface}
Taigman, Y., Yang, M., Ranzato, M., Wolf, L.:
\newblock Deepface: Closing the gap to human-level performance in face
  verification.
\newblock In: Conference on Computer Vision and Pattern Recognition (CVPR).
  (2014)

\bibitem{sun2015deepid3}
Sun, Y., Liang, D., Wang, X., Tang, X.:
\newblock Deepid3: Face recognition with very deep neural networks.
\newblock arXiv preprint arXiv:1502.00873 (2015)

\bibitem{wen2016discriminative}
Wen, Y., Zhang, K., Li, Z., Qiao, Y.:
\newblock A discriminative feature learning approach for deep face recognition.
\newblock In: European conference on computer vision (ECCV). (2016)

\bibitem{chopra2005learning}
Chopra, S., Hadsell, R., LeCun, Y.:
\newblock Learning a similarity metric discriminatively, with application to
  face verification.
\newblock In: Conference on Computer Vision and Pattern Recognition (CVPR).
  (2005)

\bibitem{sun2015deep}
Sun, Y.:
\newblock Deep learning face representation by joint
  identification-verification.
\newblock In: Advances in neural information processing systems (NIPS). (2014)

\bibitem{ustinova2016learning}
Ustinova, E., Lempitsky, V.:
\newblock Learning deep embeddings with histogram loss.
\newblock In: Advances in neural information processing systems (NIPS). (2016)

\bibitem{han2018face}
Han, C., Shan, S., Kan, M., Wu, S., Chen, X.:
\newblock Face recognition with contrastive convolution.
\newblock In: European Conference on Computer Vision (ECCV). (2018)

\bibitem{schroff2015facenet}
Schroff, F., Kalenichenko, D., Philbin, J.:
\newblock Facenet: A unified embedding for face recognition and clustering.
\newblock In: Conference on Computer Vision and Pattern Recognition (CVPR).
  (2015)

\bibitem{parkhi2015deep}
Parkhi, O.M., Vedaldi, A., Zisserman, A.:
\newblock Deep face recognition.
\newblock In: British Machine Vision Association (BMVC). (2015)

\bibitem{ge2018deep}
Ge, W.:
\newblock Deep metric learning with hierarchical triplet loss.
\newblock In: European conference on computer vision (ECCV). (2018)

\bibitem{zhong2019adversarial}
Zhong, Y., Deng, W.:
\newblock Adversarial learning with margin-based triplet embedding
  regularization.
\newblock In: International Conference on Computer Vision (ICCV). (2019)

\bibitem{oh2016deep}
Oh~Song, H., Xiang, Y., Jegelka, S., Savarese, S.:
\newblock Deep metric learning via lifted structured feature embedding.
\newblock In: Conference on Computer Vision and Pattern Recognition (CVPR).
  (2016)

\bibitem{rippel2015metric}
Rippel, O., Paluri, M., Dollar, P., Bourdev, L.:
\newblock Metric learning with adaptive density discrimination.
\newblock In: International Conference on Learning Representations (ICLR).
  (2015)

\bibitem{sohn2016improved}
Sohn, K.:
\newblock Improved deep metric learning with multi-class n-pair loss objective.
\newblock In: Advances in neural information processing systems (NIPS). (2016)

\bibitem{wu2017sampling}
Wu, C.Y., Manmatha, R., Smola, A.J., Krahenbuhl, P.:
\newblock Sampling matters in deep embedding learning.
\newblock In: International Conference on Computer Vision (ICCV). (2017)

\bibitem{sun2014deep}
Sun, Y., Wang, X., Tang, X.:
\newblock Deep learning face representation from predicting 10,000 classes.
\newblock In: Conference on Computer Vision and Pattern Recognition (CVPR).
  (2014)

\bibitem{liu2017sphereface}
Liu, W., Wen, Y., Yu, Z., Li, M., Raj, B., Song, L.:
\newblock Sphereface: Deep hypersphere embedding for face recognition.
\newblock In: Conference on Computer Vision and Pattern Recognition (CVPR).
  (2017)

\bibitem{wang2017normface}
Wang, F., Xiang, X., Cheng, J., Yuille, A.L.:
\newblock Normface: L2 hypersphere embedding for face verification.
\newblock In: Proceedings of the 25th ACM international conference on
  Multimedia (ACM). (2017)

\bibitem{wang2018additive}
Wang, F., Cheng, J., Liu, W., Liu, H.:
\newblock Additive margin softmax for face verification.
\newblock IEEE Signal Processing Letters (2018)

\bibitem{wang2018cosface}
Wang, H., Wang, Y., Zhou, Z., Ji, X., Gong, D., Zhou, J., Li, Z., Liu, W.:
\newblock Cosface: Large margin cosine loss for deep face recognition.
\newblock In: Conference on Computer Vision and Pattern Recognition (CVPR).
  (2018)

\bibitem{sun2020circle}
Sun, Y., Cheng, C., Zhang, Y., Zhang, C., Zheng, L., Wang, Z., Wei, Y.:
\newblock Circle loss: A unified perspective of pair similarity optimization.
\newblock In: Conference on Computer Vision and Pattern Recognition (CVPR).
  (2020)

\bibitem{deng2019arcface}
Deng, J., Guo, J., Xue, N., Zafeiriou, S.:
\newblock Arcface: Additive angular margin loss for deep face recognition.
\newblock In: Conference on Computer Vision and Pattern Recognition (CVPR).
  (2019)

\bibitem{zheng2018ring}
Zheng, Y., Pal, D.K., Savvides, M.:
\newblock Ring loss: Convex feature normalization for face recognition.
\newblock In: Conference on Computer Vision and Pattern Recognition (CVPR).
  (2018)

\bibitem{huang2020curricularface}
Huang, Y., Wang, Y., Tai, Y., Liu, X., Shen, P., Li, S., Li, J., Huang, F.:
\newblock Curricularface: adaptive curriculum learning loss for deep face
  recognition.
\newblock In: Conference on Computer Vision and Pattern Recognition (CVPR).
  (2020)

\bibitem{meng2021magface}
Meng, Q., Zhao, S., Huang, Z., Zhou, F.:
\newblock Magface: A universal representation for face recognition and quality
  assessment.
\newblock In: Conference on Computer Vision and Pattern Recognition (CVPR).
  (2021)

\bibitem{chen2017noisy}
Chen, B., Deng, W., Du, J.:
\newblock Noisy softmax: Improving the generalization ability of dcnn via
  postponing the early softmax saturation.
\newblock In: Conference on Computer Vision and Pattern Recognition (CVPR).
  (2017)

\bibitem{kim2020broadface}
Kim, Y., Park, W., Shin, J.:
\newblock Broadface: Looking at tens of thousands of people at once for face
  recognition.
\newblock In: European Conference on Computer Vision, Springer (2020)

\bibitem{deng2021variational}
Deng, J., Guo, J., Yang, J., Lattas, A., Zafeiriou, S.:
\newblock Variational prototype learning for deep face recognition.
\newblock In: Proceedings of the IEEE/CVF Conference on Computer Vision and
  Pattern Recognition. (2021)

\bibitem{scheirer2012toward}
Scheirer, W.J., de~Rezende~Rocha, A., Sapkota, A., Boult, T.E.:
\newblock Toward open set recognition.
\newblock IEEE transactions on pattern analysis and machine intelligence (2012)

\bibitem{geng2020recent}
Geng, C., Huang, S.j., Chen, S.:
\newblock Recent advances in open set recognition: A survey.
\newblock IEEE transactions on pattern analysis and machine intelligence (2020)

\bibitem{ge2017generative}
Ge, Z., Demyanov, S., Chen, Z., Garnavi, R.:
\newblock Generative openmax for multi-class open set classification.
\newblock arXiv preprint arXiv:1707.07418 (2017)

\bibitem{yoshihashi2019classification}
Yoshihashi, R., Shao, W., Kawakami, R., You, S., Iida, M., Naemura, T.:
\newblock Classification-reconstruction learning for open-set recognition.
\newblock In: Conference on Computer Vision and Pattern Recognition (CVPR).
  (2019)

\bibitem{zhang2019adacos}
Zhang, X., Zhao, R., Qiao, Y., Wang, X., Li, H.:
\newblock Adacos: Adaptively scaling cosine logits for effectively learning
  deep face representations.
\newblock In: Conference on Computer Vision and Pattern Recognition (CVPR).
  (2019)

\bibitem{liu2019adaptiveface}
Liu, H., Zhu, X., Lei, Z., Li, S.Z.:
\newblock Adaptiveface: Adaptive margin and sampling for face recognition.
\newblock In: Conference on Computer Vision and Pattern Recognition (CVPR).
  (2019)

\bibitem{lu2019sampling}
Lu, J., Xu, C., Zhang, W., Duan, L.Y., Mei, T.:
\newblock Sampling wisely: Deep image embedding by top-k precision
  optimization.
\newblock In: Proceedings of the IEEE/CVF International Conference on Computer
  Vision. (2019)

\bibitem{liu2021dam}
Liu, J., Wu, Y., Wu, Y., Li, C., Hu, X., Liang, D., Wang, M.:
\newblock Dam: Discrepancy alignment metric for face recognition.
\newblock In: Proceedings of the IEEE/CVF International Conference on Computer
  Vision. (2021)

\bibitem{deng2019lightweight}
Deng, J., Guo, J., Zhang, D., Deng, Y., Lu, X., Shi, S.:
\newblock Lightweight face recognition challenge.
\newblock In: International Conference on Computer Vision Workshops (ICCVW).
  (2019)

\bibitem{guo2016ms}
Guo, Y., Zhang, L., Hu, Y., He, X., Gao, J.:
\newblock Ms-celeb-1m: A dataset and benchmark for large-scale face
  recognition.
\newblock In: European conference on computer vision (ECCV). (2016)

\bibitem{huang2008labeled}
Huang, G.B., Mattar, M., Berg, T., Learned-Miller, E.:
\newblock Labeled faces in the wild: A database forstudying face recognition in
  unconstrained environments.
\newblock In: Workshop on faces in'Real-Life'Images: detection, alignment, and
  recognition. (2008)

\bibitem{sengupta2016frontal}
Sengupta, S., Chen, J.C., Castillo, C., Patel, V.M., Chellappa, R., Jacobs,
  D.W.:
\newblock Frontal to profile face verification in the wild.
\newblock In: IEEE Winter Conference on Applications of Computer Vision (WACV).
  (2016)

\bibitem{zheng2018cross}
Zheng, T., Deng, W.:
\newblock Cross-pose lfw: A database for studying cross-pose face recognition
  in unconstrained environments.
\newblock Beijing University of Posts and Telecommunications, Tech. Rep (2018)

\bibitem{moschoglou2017agedb}
Moschoglou, S., Papaioannou, A., Sagonas, C., Deng, J., Kotsia, I., Zafeiriou,
  S.:
\newblock Agedb: the first manually collected, in-the-wild age database.
\newblock In: Conference on Computer Vision and Pattern Recognition Workshops
  (CVPRW). (2017)

\bibitem{zheng2017cross}
Zheng, T., Deng, W., Hu, J.:
\newblock Cross-age lfw: A database for studying cross-age face recognition in
  unconstrained environments.
\newblock arXiv preprint arXiv:1708.08197 (2017)

\bibitem{whitelam2017iarpa}
Whitelam, C., Taborsky, E., Blanton, A., Maze, B., Adams, J., Miller, T.,
  Kalka, N., Jain, A.K., Duncan, J.A., Allen, K.,  et~al.:
\newblock Iarpa janus benchmark-b face dataset.
\newblock In: Conference on Computer Vision and Pattern Recognition workshops
  (CVPRW). (2017)

\bibitem{maze2018iarpa}
Maze, B., Adams, J., Duncan, J.A., Kalka, N., Miller, T., Otto, C., Jain, A.K.,
  Niggel, W.T., Anderson, J., Cheney, J.,  et~al.:
\newblock Iarpa janus benchmark-c: Face dataset and protocol.
\newblock In: International Conference on Biometrics (ICB). (2018)

\bibitem{he2016deep}
He, K., Zhang, X., Ren, S., Sun, J.:
\newblock Deep residual learning for image recognition.
\newblock In: Conference on Computer Vision and Pattern Recognition (CVPR).
  (2016)

\bibitem{deng2020sub}
Deng, J., Guo, J., Liu, T., Gong, M., Zafeiriou, S.:
\newblock Sub-center arcface: Boosting face recognition by large-scale noisy
  web faces.
\newblock In: European Conference on Computer Vision, Springer (2020)

\bibitem{shi2020towards}
Shi, Y., Yu, X., Sohn, K., Chandraker, M., Jain, A.K.:
\newblock Towards universal representation learning for deep face recognition.
\newblock In: Proceedings of the IEEE/CVF Conference on Computer Vision and
  Pattern Recognition. (2020)  6817--6826

\bibitem{li2021spherical}
Li, S., Xu, J., Xu, X., Shen, P., Li, S., Hooi, B.:
\newblock Spherical confidence learning for face recognition.
\newblock In: Proceedings of the IEEE/CVF Conference on Computer Vision and
  Pattern Recognition. (2021)

\bibitem{kim2020groupface}
Kim, Y., Park, W., Roh, M.C., Shin, J.:
\newblock Groupface: Learning latent groups and constructing group-based
  representations for face recognition.
\newblock In: Proceedings of the IEEE/CVF Conference on Computer Vision and
  Pattern Recognition. (2020)

\end{thebibliography}
\end{document}


\pagestyle{headings}
\mainmatter

\def\ACCV22SubNumber{144}  

\title{Supplementary Meterial of GB-CosFace} 
\titlerunning{ACCV-22 submission ID \ACCV22SubNumber}
\authorrunning{ACCV-22 submission ID \ACCV22SubNumber}

\author{Anonymous ACCV 2022 submission}
\institute{Paper ID \ACCV22SubNumber}

\maketitle

\section{Details of Softmax-based Loss Derivation}

For the objective design of face recognition, following the multi-classification pipeline, the direct idea is to constrain the target score larger than the maximum non-target score. For example, given a training sample and its label y, the base objective is as follows:

\begin{equation}
    \displaystyle
      \mathcal{O}_{base}=ReLU(max({p_i})-p_y, 0)
    \label{Eq.Obase}
\end{equation}%
Where \(p_y\) is the target score and \(p_i\) is the non-target score. ReLU is equivalent to
the function \(max(\cdot, 0)\), which is added to avoid overfitting. 

For face recognition, as has been discussed in \textbf{Section 2} in the main text, the predicted score can be represented as the cosine of the face feature vector and the prototype, and usually, the margin is introduced for stricter constraints. Therefore, we can get the following training objective for face recognition:

\begin{equation}
    \displaystyle
       \mathcal{O}_{S}= ReLU(max(cos\theta_i )-(cos(\theta_y+m_\theta  )-m_p))
    \label{Eq.OS}
\end{equation}%

However, generally, we use the softmax-based loss for training, which is the smooth form of \(\mathcal{O}_{S}\) based on the following equations:

\begin{equation}
    \displaystyle
       ReLU(x)=\lim_{s \to \infty} \frac{1}{s}log(1+e^{sx}) 
    \label{Eq.lim1}
\end{equation}%

\begin{equation}
    \displaystyle
       max(p_i)=\lim_{s \to \infty} \frac{1}{s}log\sum_{i=1}^{N} e^{sp_i}
    \label{Eq.lim2}
\end{equation}%

The general softmax-based loss is as follows.

\begin{equation}
    \displaystyle
    {\mathcal{L}_S=-log\frac{e^{s{(cos(\theta _y +m_\theta )}-m_p)}}{ {\textstyle e^{s(cos(\theta _y +m_\theta )-m_p)}+\sum_{i}e^{s{cos_{\theta i}}}}}}
    \label{Eq.L1} 
\end{equation}%

Substituting \textbf{Equ.} \ref{Eq.lim1} and \textbf{Equ.} \ref{Eq.lim2} into \textbf{Equ.} \ref{Eq.OS}, we can reach the following equation.

\begin{equation}
    \displaystyle
      \begin{aligned}
      \mathcal{O}_{S}&= ReLU(max(cos\theta_i )-(cos(\theta_y+m_\theta  )-m_p)) \\
        &= \lim_{s \to +\infty}\frac{1}{s}log(1+e^{(log\sum_{i=i,i\ne y}^{N}e^{scos\theta_i})-s\cdot (cos(\theta _y+m_\theta )-m_p)  }) \\
        &= \lim_{s \to +\infty}\frac{1}{s}log(1+{\frac{\sum_{i=1,i\ne y}^{n}e^{scos\theta_i}}{e^{s\cdot (cos(\theta _y+m_\theta )-m_p}}} ) \\
        &= \lim_{s \to +\infty}-\frac{1}{s}log\frac{e^{s(cos(\theta_y+m_\theta)-m_p)}}{e^{s(cos(\theta_y+m_\theta)-m_p)}+\sum_{i=1,i\ne y}^{n}e^{scos\theta_i}} \\ 
        &= \lim_{s \to +\infty}\frac{1}{s} \mathcal{L}_S
    \end{aligned} 
    \label{Eq.smooth}
\end{equation}%

In \textbf{Equ.} \ref{Eq.smooth}, if we take a fixed value of \(s\) rather than the limit of positive infinity, the form of the softmax-based loss can be reached.

\section{Details of GB-CosFace Antetype Derivation}

As has been discussed in \textbf{Section 3.1} in the main text, the original objective is as follows.

\begin{equation}
    \displaystyle
    \left\{\begin{array}{l}
    \mathcal{O}_{T}=ReLU(p_v-(p_y-m))
    \\
    \mathcal{O}_{N}=ReLU(max(p_i)-(p_v-m))
    
    \end{array}\right.
    \label{Eq.etn}
\end{equation}%

Similar to \textbf{Equ.} \ref{Eq.smooth}, we substitute \textbf{Equ.} \ref{Eq.lim1} and \textbf{Equ.} \ref{Eq.lim2} into \textbf{Equ.} \ref{Eq.etn} as follows.
\begin{equation}
    \displaystyle
    \begin{aligned}
      \mathcal{O}_{T}&= ReLU(p_v-(p_y-m)) \\
        &= \lim_{s \to +\infty}\frac{1}{s}log(1+ e^{s(p_v-(p_y-m))}) \\
        &= \lim_{s \to +\infty}\frac{1}{s}log(1+ \frac{e^{sp_v}}{e^{s(p_y-m)} }) \\
       &= \lim_{s \to +\infty}-\frac{1}{s}log(\frac{e^{s(p_y-m)}}{e^{s(p_y-m)}+e^{sp_v}} ) 
    \end{aligned} 
    \label{Eq.etn}
\end{equation}%

\begin{equation}
    \displaystyle
    \begin{aligned}
      \mathcal{O}_{N}&= ReLU(max(p_i)-(p_v-m)) \\
        &= \lim_{s \to +\infty}\frac{1}{s}log(1+ e^{log\sum_{i}^{}e^{sp_i} -s(p_v-m)}) \\
        &= \lim_{s \to +\infty}\frac{1}{s}log(1+ \frac{\sum_{i}^{}e^{sp_i}}{e^{s(p_y-m)}}) \\
       &= \lim_{s \to +\infty}-\frac{1}{s}log(\frac{e^{s(p_y-m)}}{e^{s(p_y-m)}+{\sum_{i}^{}e^{sp_i}}} ) 
    \end{aligned} 
    \label{Eq.etn}
\end{equation}%

Taking parameter \(s\) as a fixed value rather than positive infinity, we can reach the following loss.

\begin{equation}
    \displaystyle
    \left\{\begin{array}{l}
    \mathcal{L}_{T1}=-log\frac{e^{s(p_y-m)}}{e^{s(p_y-m)}+e^{sp_v}} 
    \\
    \mathcal{L}_{N1}=-log\frac{e^{s(p_v-m)}}{e^{s(p_v-m)}+ {\textstyle \sum_{i}} e^{sp_i}} 
    \end{array}\right.
    \label{Eq.base}
\end{equation}%

\section{Proof of the Compatibility with CosFace}
The final loss is as follows.

\begin{equation}
    \displaystyle
    \begin{aligned}
      \mathcal{L}_{GB-CosFace} &= -\frac{1}{2} log\frac{e^{2s(p_y-m)}}{e^{2s(p_y-m)}+e^{2sp_v}}\\
        &\quad-\frac{1}{2}log\frac{e^{2s(p_v-m)}}{e^{2s(p_v-m)}+  e^{2sp_n}} \\
    \end{aligned} 
    \label{Eq.final}
\end{equation}%
Where \(p_{n}=\frac{1}{s}log {\textstyle \sum_{i}}  e^{sp_{i}}\). The parameter \(p_v\) is determined by the following equation.

\begin{equation}
    \displaystyle
    {p_v=\alpha p_{vg}+(1-\alpha)\hat{p}_v}
    \label{Eq.alpha}
\end{equation}%

Where \(\hat{p}_v=(p_y+p_n )/2\), and \(p_{vg}\)is the global boundary.

The proposed GB-CosFace has the following property.

\textbf{Property 3} For \(\alpha=0\), GB-CosFace with the scale parameter \(s\) and the margin parameter \(m\) is equivalent to CosFace with scale parameter \(s\) and the margin parameter \(2m\).

\(Proof.\) 
To prove the compatibility with CosFace\cite{wang2018additive,wang2018cosface}, we calculate the gradient for the target score \(\mathcal{G}_{CosFace}^y\) and the gradient for the non-target score \(\mathcal{G}_{CosFace}^i\) under CosFace framework.

\begin{equation}
    \displaystyle
    \mathcal{G}_{CosFace}^y=-\frac{s\cdot  {\textstyle \sum_{i}} e^{sp_i}}{e^{s(p_y-m)}+\sum_{i} e^{sp_i}} 
    \label{Eq.cos_gy}
\end{equation}%

\begin{equation}
    \displaystyle
    \mathcal{G}_{CosFace}^i=\frac{s\cdot e^{sp_i}}{e^{s(p_y-m)}+\sum_{i} e^{sp_i}}   
    \label{Eq.cos_gi}
\end{equation}%

For GB-CosFace with \(\alpha=0\), \(p_v=(p_y+p_n )/2\). The gradient for the target score \(\mathcal{G}_{GB}^y\) is as follows. Note that \(p_v\) is a detached parameter which does not require gradients.

\begin{equation}
    \displaystyle
    \begin{aligned}
      \mathcal{G}_{GB}^y&=\frac{-s\cdot e^{2sp_v}}{e^{2s(p_y-m)}+e^{2sp_v}}  \\
        &=\frac{-s\cdot e^{2s((p_y+p_n)/2)}}{e^{2s(p_y-m)}+e^{2s((p_y+p_n)/2)}}\\
        &=\frac{-s\cdot e^{s(p_y+p_n)}}{e^{2s(p_y-m)}+e^{s(p_y+p_n)}}\\
        &=\frac{-s\cdot e^{sp_n}}{e^{s(p_y-2m)}+e^{sp_n}}\\
        &=-\frac{s\cdot  {\textstyle \sum_{i}} e^{sp_i}}{e^{s(p_y-2m)}+\sum_{i} e^{sp_i}}
        =\mathcal{G}_{CosFace}^y
    \end{aligned}  
    \label{Eq.gb_y}
\end{equation}%

The gradient for the non-target score \(\mathcal{G}_{GB}^i\) is as follows.

\begin{equation}
    \displaystyle
    \begin{aligned}
      \mathcal{G}_{GB}^i&=\frac{\partial \mathcal{L}_{GB-CosFace}}{\partial p_n}\cdot \frac{\partial p_n}{\partial p_i}\\
        &=\frac{s\cdot e^{2sp_n}}{e^{2s(p_v-m)}+e^{2sp_n}}\cdot \frac{e^{sp_i}}{ {\textstyle \sum_{i}e^{sp_i}} }\\
        &=\frac{s\cdot e^{2sp_n}}{e^{s((p_y+p_n)-2m)}+e^{2sp_n}}\cdot \frac{e^{sp_i}}{ {\textstyle \sum_{i}e^{sp_i}} }\\
        &=\frac{s\cdot e^{sp_n}}{e^{s(p_y-2m)}+e^{sp_n}}\cdot \frac{e^{sp_i}}{ {\textstyle \sum_{i}e^{sp_i}} }\\
        &=\frac{s\cdot \sum_{i}e^{sp_i}}{e^{s(p_y-2m)}+e^{sp_n}}\cdot \frac{e^{sp_i}}{ {\textstyle \sum_{i}e^{sp_i}} }\\
        &=\frac{s\cdot e^{sp_i}}{e^{s(p_y-2m)}+\sum_{i}e^{sp_i}}
        =\mathcal{G}_{CosFace}^i
    \end{aligned} 
    \label{Eq.gb_y}
\end{equation}%

\section{Effect of the update rate \(\gamma\)}
We expect to calculate \(\hat{p}_v\) in \textbf{Equ.} \ref{Eq.alpha} for each sample in the data set and get the mean value as the threshold \(p_v\). The momentum update strategy is applied to update the mean of \(\hat{p}_v\) for each batch.
\begin{equation}
    \displaystyle
    {p_{vg}=(1-\gamma)p_{vg}+\gamma p_{vb}}
    \label{Eq.vg}
\end{equation}%
Where \(\gamma \in [0,1]\) is the update rate, \(p_{vb}\) is the mean of \(p_{v}\) in a batch. According to our experience, as long as this \(\gamma\) is set in a small reasonable range, it will not have much impact on our experimental results. To evaluate the effectiveness of the update rate \(\gamma\), we empirically set \(s=32\), \(m=0.16\), \(\alpha=0.15\) and focus on the setting of \(\gamma\). As can be seen from the \textbf{Table} \ref{tab:gamma}, when the \(\gamma\) value is set relatively small, around 0.01, the performance of the model can maintain relatively good results. Therefore, we empirically set \(\gamma\) to 0.01 is reasonable.
\begin{table}[hbpt]
\begin{center}

    \caption{The results of the proposed GB-CosFace under different settings of update rate \(\gamma\).}
    \label{tab:gamma}
    \setlength{\tabcolsep}{6pt}{
    \centering
    \scalebox{1}{
    \begin{tabular}{l|c}
    \hline 
    {\centering Settings} &IJB-C(TAR)\\
    \hline
    FAR=1e-4, R50, \(\alpha\)=0.15, \(\gamma\)=0.001 & 96.25\\
    FAR=1e-4, R50, \(\alpha\)=0.15, \(\gamma\)=0.003 & 96.27\\
    FAR=1e-4, R50, \(\alpha\)=0.15, \(\gamma\)=0.005 & 96.19\\
    FAR=1e-4, R50, \(\alpha\)=0.15, \(\gamma\)=0.01 & 96.24\\
    FAR=1e-4, R50, \(\alpha\)=0.15, \(\gamma\)=0.05 & 96.33\\
    \hline
    \end{tabular}}
    }
    \end{center}
    \vspace{-0.5cm}
\end{table}

\clearpage


\bibliographystyle{splncs}
\bibliography{egbib}
